%% file: GP_with_banded_ops.tex
\documentclass[twoside]{article}

\usepackage[accepted]{aistats2019}

\usepackage[utf8]{inputenc} %
\usepackage[T1]{fontenc}    %
\PassOptionsToPackage{hyphens}{url}\usepackage[hidelinks]{hyperref}       %
\usepackage{booktabs}       %
\usepackage{amsfonts}       %
\usepackage{nicefrac}       %
\usepackage{microtype}      %
\usepackage{array}%
\usepackage{float}
\usepackage{xfrac}

\newcommand{\spc}[2][c]{%
  \begin{tabular}[#1]{@{}c@{}}#2\end{tabular}}

\usepackage[round]{natbib}

\usepackage{amsmath}
\usepackage{dsfont}
\usepackage{xcolor}
\usepackage{algorithm} 
\usepackage{algorithmic}  
\usepackage[algo2e]{algorithm2e}
\usepackage{graphicx}
\usepackage{subcaption}
\newcommand{\todo}[1]{}
\renewcommand{\todo}[1]{{\color{teal} \textit{TODO: {#1}}}}
\newcommand{\james}[1]{}
\renewcommand{\james}[1]{{\color{green} \textit{James: {#1}}}}
\newcommand{\tmp}[1]{}
\renewcommand{\tmp}[1]{{\color{gray} \textit{Temporary: {#1}}}}
\newcommand{\tocheck}[1]{}
\renewcommand{\tocheck}[1]{{\color{blue} #1}}

\newcommand{\cD}{\mathcal{D}}
\newcommand{\cL}{\mathcal{L}}
\newcommand{\NN}{\mathcal{N}}
\newcommand{\EE}{\mathbb{E}}
\newcommand{\RR}{\mathbb{R}}
\newcommand*{\KL}[2]{\ensuremath{\operatorname{KL}[#1 \,\|\, #2]}}

\newcommand{\PP}{\mathds{P}}
\newcommand{\Ss}{\mathds{S}}

\newcommand{\OO}{\mathds{O}}

\newcommand{\N}{N}
\newcommand{\n}{n}
\renewcommand{\top}{T}

\newcommand{\opS}{\mathds{S}}
\newcommand{\opI}{\mathds{I}}
\newcommand{\opO}{\mathds{O}}
\newcommand{\opP}{\mathds{P}}
\newcommand{\opC}{\mathds{C}}

\title{Banded Matrix Operators for Gaussian Markov Models in the Automatic Differentiation Era}

\begin{document}

\twocolumn[

\aistatstitle{Banded Matrix Operators for Gaussian Markov Models in the Automatic Differentiation Era}

\aistatsauthor{Nicolas Durrande, Vincent Adam, Lucas Bordeaux, Stefanos Eleftheriadis and James Hensman}

\aistatsaddress{PROWLER.io, Cambridge, UK }]

\begin{abstract}
    Banded matrices can be used as precision matrices in several models including linear state-space models, some Gaussian processes, 
    and Gaussian Markov random fields. 
    The aim of the paper is to make modern inference methods (such as variational inference or gradient-based 
    sampling) available for Gaussian models with banded precision. 
    We show that this can efficiently be achieved by equipping an automatic differentiation framework, such as TensorFlow or PyTorch,
    with some linear algebra operators dedicated to banded matrices. 
	This paper studies the algorithmic aspects of the required operators, details their reverse-mode derivatives, 
	and show that their complexity is linear in the number of observations.

\end{abstract}

\input{intro.tex}

\input{background.tex}

\input{inference.tex}

\input{ops.tex}

\input{experiments.tex}

\bibliographystyle{plainnat}
\bibliography{GP_with_banded_ops} 

\cleardoublepage
\onecolumn
\section*{Supplementary materials}
\appendix
\input{appendix_derivations.tex}

\input{appendix.tex}
\input{appendix_algo_pseudocode.tex}

\end{document}

%% file: intro.tex
\section{Introduction}

Gaussian process (GP) modelling is a popular framework for predicting the value
of a (latent) function $f$ given a limited set of input/output observation
tuples. It encapsulates several common methods such as linear regression,
smoothing splines and the reproducing kernel Hilbert space approximation
\citep{rasmussen2006gaussian}. The popularity of this framework can be explained by 
its efficiency when little data is available~\citep{sacks1989design},
the existence of an analytical solution for the posterior when the likelihood is Gaussian,
and the control over the prior that is offered by the choice the covariance function (i.e.\ the kernel).

Two practical limitations of GP models are that algorithms for computing the posterior distribution typically
scale in $\mathcal{O}(n^2)$ space and $\mathcal{O}(n^3)$ time where $n$ is the number of observations, and that the
posterior distribution is not tractable when the likelihood is not conjugate.
These two limitations have been thoroughly studied over the past decades and
several approaches have been proposed to overcome them. The most popular method
for reducing computational complexity is the sparse GP framework
\citep[][]{candela2005, titsias2009}, where computations are focussed on a set
of ``inducing variables'', allowing a trade-off between computational
requirements and the accuracy of the approximation. To cope with non-conjugacy
in these models, several approximation methods such as the Laplace approximation,
variational inference (VI) or expectation propagation have been proposed to approximate
 non-Gaussian posteriors by Gaussian distributions~\citep{nickisch2008approximations}.

Another angle to tackle the complexity inherent to GP models is to choose a class
of covariance functions that lead to particular structures that 
can be exploited for storage and/or computational gain. 
Although several efficient methods are based on structured \emph{covariance} matrices $K$ \citep{gneiting2002compactly,nickson2015blitzkriging,wilson2015kernel}, state space models ~\citep[SSM, see][]{sarkka2013bayesian}---including the iconic Kalman filter~\citep{kalman1960}---and Gaussian
Markov random fields~\citep[GMRF,][]{rue2005gaussian} are using sparse structure in the 
\emph{precision} matrix $Q=K^{-1}$. We will refer to these methods using sparsity in
the precision matrix as Gaussian Markov models. Exploiting this sparsity can bring orders of magnitude speed-ups compared
to naive implementations based on covariance matrices. Furthermore, it can be proved that some
classical covariance functions have an equivalent state space representation that leads to a sparse precision~\citep[see Section~\ref{sec:background} and][]{solin2016stochastic}.

Learning the hyper-parameters of GP models parameterised by their precision can be challenging, and it is common to resort to
Markov chain Monte Carlo (MCMC) sampling~\cite{rue2005gaussian}. In this context, it is however 
recognized that typical MCMC samplers such as Metropolis Hastings or Gibbs sampling
suffer from high correlation between the latent variables \citep{rue2009approximate}.
Deterministic approximations based on the Laplace approximation have been derived, such
as the widely used integrated nested Laplace approximations \citep{rue2009approximate}.
Our aim however is to provide general inference methods that can be applied to a broader class of models (e.g.\ beyond the scope of the 
classical combination of a Gaussian latent function with an associated likelihood).

The main contribution of this article is to show that the limitations of current inference methods for precision-based models can be overcome by implementing a small set of low level linear algebra operators dedicated to banded matrices and their derivatives in an automatic differentiation framework such as TensorFlow \citep{abadi2016tensorflow}.
We propose a general framework that allows us to perform marginal likelihood estimation for models with conjugate likelihoods, Hamiltonian Monte Carlo (HMC) and VI in linear complexity both in time and space in the non-conjugate setting. 

Most inference and learning algorithms in Gaussian models involve a small set of linear algebraic operations, such as matrix product, Cholesky factorisation or triangular solve.
For general Gaussian models, efficient implementations of these operations and their derivatives have been proposed \citep{murray2016differentiation, seeger2017auto, giles2008collected}. Tailored primitives have been designed for SSMs \citep{nickisch2018state, grigorievskiy2016parallelizable}. These however lack derivations of their reverse-mode differentiation, which prevents their use in automatic differentiation libraries. With this paper we fill this gap by introducing a set of linear algebra operations for Gaussian models with banded precisions. Compared to the dense case (i.e.\ precisions without band structure), our framework also includes dedicated algorithms to compute subsets of the inverses of sparse matrices \citep{takahashi1973formation, zammit2018sparse}.

In this paper, we revisit and develop inference and learning algorithms in GP models with banded precisions following the wide adoption of end-to-end training of generative models using automatic differentiation.
The paper is organised as follows: Section~\ref{sec:background} gives some context and background on precision matrices with a focus on the banded case. In Section~\ref{sec:inference} we survey inference and learning algorithms for Gaussian models with banded precision and identify the basic linear algebraic operations they require. In Section~\ref{sec:op} we describe how these operations (and their derivatives) can be efficiently implemented. In Section~\ref{sec:experiments} 
we show on two experiments based on SSMs and GMRFs that the proposed framework scales to large problems and is proven to be attractive for real-world scenarios.

%% file: background.tex
\section{Background on banded precision matrices}%
\label{sec:background}

For a Gaussian random vector $g$ of length $N$ with covariance matrix $K$, the element $K_{i, j}$ of
the covariance matrix corresponds to the covariance between $g_i$ and $g_j$.
Elements of covariance matrices thus correspond to marginal distributions,
all other variables being marginalised out. The interpretation of the elements of
a precision matrix $Q = K^{-1}$ is not as straightforward, but it is still possible:
$Q_{i,j}$ is a function of the conditional distribution of $g_i$, $g_j$ given all other variables.
More precisely, let $I$ be  a subset of $\{1, \ldots, N\}$ and let $Q_{I, I}$ and $g_I$ be the restriction of $Q$ and
$g$ to the indices in $I$, then $Q_{I, I} = (\mathrm{cov}(g_I, g_I \mid g_k,\ k \notin I))^{-1}$. 
Taking $I=\{i\}$ shows that $Q_{i, i} = (\mathrm{var}(g_i \mid g_j,\ j \neq i))^{-1}$: 
contrary to the covariance case where extracting a sub-covariance amounts to marginalisation, a
sub-precision corresponds to the inverse of a conditional covariance.
Similarly, choosing $I=\{i, j\}$, one can show that conditional independence between $g_i$ and $g_j$
(given all other variables) implies $Q_{i,j}=0$. This means that random vectors with conditional independences
will lead to sparse precision matrices.

Banded matrices are sparse matrices that only have non-zero values within a small “band”
around their diagonal. The lower and upper bandwidths of a banded matrix $B$ are defined as the 
smaller integers $l_l$ and $l_u$ such that $i + l_u < j < i - l_l$ implies $B_{i,j}=0$.  For example, 
a tridiagonal matrix has $l_l = l_u = 1$. The bandwidth of a matrix is $l = \max(l_l, l_u)$.

In one dimension, a typical example of conditional independence resulting in banded precisions is given by random vectors with
the Markov property. For example if $g$ corresponds to the evaluations of a one-dimensional GP $f$ with
Brownian or Mat\'ern\sfrac{1}{2} covariance at increasing input locations
($g_i = f(x_i)$ with $x_i < x_j$ for $i<j$), then $Q$ is banded with lower
and upper bandwidths equal to one.
In a similar fashion, one dimensional GPs with higher order Mat\'ern kernels can also lead to banded
precision matrices, but it is necessary to augment the state-space dimension by adding some derivatives.
For example, the vector $g = (f(x_0), f'(x_0), f(x_1), f'(x_1), \ldots, f'(x_n))$ where $f$ is
a GP with Mat\'ern\sfrac{3}{2} covariance will result in a precision with lower (and upper)
bandwidth equal to three~\citep{grigorievskiy2016parallelizable}.

Other kernels such as the squared exponential do not result in banded precision matrices. It is however
possible to find a good approximation of the covariance such that the precision is banded as
discussed by \citet{sarkka2014convergence}. A final one-dimensional example resulting in banded precisions
are autoregressive models~\citep{jones1981fitting}.  

In higher dimensions, when the GP input is $x \in \mathds{R}^d$, there is no direct
equivalent of the Markov property. The classical approach is to consider a set 
$V= \{x_1, \ldots, x_N \}$ of points $x_i \in \mathds{R}^d$
and to define a set of undirected edges $E$ between these points to obtain a graph structure.
Now, let $g$ be a Gaussian random vector corresponding to the evaluation of a GP $h$ indexed by the nodes of
the graph $g_i = h(x_i)$. It is then possible to have an equivalent of the Markov
property where, given the values of $g$ at the neighbouring nodes $\{k,\ (i, k) \in E\}$, $g_i$ is independent of the rest of the graph.
Assuming that, given all other entries, $g_i$ and $g_j$ are independent is equivalent to considering
a precision matrix $Q_{i,j}$ satisfying $Q_{i,j}=0$ if $(i,j) \notin E$. One example 
is the Laplacian precision $Q = D - A$, where $D$ is a diagonal matrix with $D_{i, i} = \mathrm{degree}(i)$ 
called the degree matrix, 
and $A$ is the adjacency matrix: $A_{i, j} = 1$ if $(i, j) \in E$ and
$0$ otherwise~\citep{belkin2004regularization}. Although this leads to a sparse precision $Q$, the associated bandwidth
depends on the ordering of the nodes and it is possible to use heuristics such as the Cuthill McKee
algorithm to find a node ordering associated to a thin bandwidth~\citep{rue2005gaussian}.

%% file: inference.tex
\section{Fast inference with banded precisions }
\label{sec:inference}
In this section we look at three inference techniques and investigate the banded matrix operations that are required for each case. 
Let $X = [x_1, \dots x_\n]^T$ with $x_i \in \cD \subseteq \mathds{R}^d$ and $Y=[y_1, \dots, y_\n]^T$ with $y_i \in \mathds{R}$
denote matrices corresponding to input and output values of the data. We consider the following type of models:
a latent function $f$ is defined over $\cD$; the prior on the latent function is parameterised by $\theta$; given the latent function, an observation model provides a likelihood that factorises as $p(Y | f) = \prod_{i=1}^\n p(y_i | F_i)$, with $F = f(X)$.

\subsection{Marginal likelihood computation in tractable problems}
\label{subsec:loglik}

In the case where the likelihood is Gaussian with variance $\tau^2$, the common approach for estimating
the model parameters $\theta$ is to maximise the marginal likelihood $p(Y|\theta)$. This requires computing
the prior distribution of the latent function at locations where observations are provided. This is
straightforward when the latent function is parameterised by its covariance, as in a Gaussian process model, but it scales cubically with the number of observations. \citet{grigorievskiy2016parallelizable} show that it is possible to do this computation efficiently for an SSM with banded precision. 
The formulation of the SSM allows them to compute the precision matrix for an arbitrary subset of the total input locations.
In the case of GMRF, the precision matrix $Q$ is specified for the all the nodes of the graph, it thus has a size $N \times N$ even if the $n$ observations are only associated to a subset of nodes.
We show below that the approach of \citet{grigorievskiy2016parallelizable} can be generalised to this case. The covariance matrix of $f(X)$ is $K = E Q^{-1} E^T$ where $E$ is an $\n \times \N$ matrix of 0 and 1 that selects the appropriate rows of $Q^{-1}$. Using the matrix inversion and the matrix determinant lemma, we obtain:
\begin{align}
    \nonumber\log\,  p(Y| &\theta) = -\frac{\n}{2} \log(2 \pi) - \frac12 \log |E Q^{-1} E^T + \tau^2 I| \\
\nonumber & \qquad \qquad - \frac12 Y^T (E Q^{-1} E^T + \tau^2 I)^{-1} Y\\
\nonumber    &= -\frac{\n}{2} \log(2 \pi) - \frac12 \log |Q  + \tau^{-2} E^T E| \\
\nonumber & \qquad + \frac12 \log |Q|  - \frac12 \log| \tau^{2} I| - \frac{1}{2 \tau^2} Y^T Y \\
    &\qquad + \frac{1}{2 \tau^4} Y^T E (Q  + \tau^{-2} E^TE)^{-1}E^T Y.
\end{align}
The definition of $E$ implies that $E^TE$ is a diagonal matrix so $Q + \tau^{-2}E^TE$ has the same bandwith as $Q$. The Cholesky factors of banded positive-definite matrices are lower triangular banded matrices that can be computed efficiently (see Section~\ref{sec:op}).
We thus introduce $LL^T = (Q  + \tau^{-2} E^T E)$ and $L_Q^{}L_Q^T = Q$ to obtain the following expression of the marginal likelihood:
\begin{align}
    \nonumber    \log\,p(Y| \theta) &=  -\frac{\n}{2} \log(2 \pi) - \log |L| + \log |L_Q| \\
    \nonumber              & \qquad - \frac{n}{2} \log \tau^{2} - \frac{1}{2 \tau^2} Y^T Y \\
              & \qquad + \frac{1}{2 \tau^4} Y^T E L^{-T} L^{-1} E^T Y.
\label{eq:marg_lik}
\end{align}
As a consequence, the computation of the marginal likelihood requires the efficient computation of the Cholesky factorisation of banded matrices and the efficient solution of linear systems with banded triangular matrices such as $L^{-1} (E^T Y)$ in Eq.~\ref{eq:marg_lik}.

\subsection{Gradient-based MCMC}
\label{subsec:hmc}

Asymptotically exact inference in non-conjugate models is achievable through MCMC sampling. Among available algorithms, the most
empirically effective are HMC \citep[see
e.g.,][]{neal2011mcmc} and its variants \citep{hoffman2014no}. These samplers
require the log joint density of the latent variables and the data, $\log \, p(F,Y | \theta)$,
as well as its derivatives with respect to the latent variables. 
Here we investigate which operations involving banded matrices are required for this purpose. 

To avoid strong correlations in the joint distribution, which reduces the
effectiveness of the sampler, whitening of the latent variables is often
employed \citep[see e.g.,][]{filippone2013comparative}. 
Let $v \sim \mathcal N (0, I)$ be a random vector of length $\N$. %
One can generate samples of $f(X)$ by computing $F=L_K v$ where $L_K$ is the Cholesky factor of the covariance matrix $K$ of $f(X)$.
When working with a precision $Q$, we can instead write $F =
L_Q^{-\top}v$, resulting in $p(F) = \mathcal N(0, Q^{-1})$ as required.

The log joint density is then:
\begin{align}
\nonumber   \log \, p(v, \theta, Y) &= \log\,p(v) +  \log\,p(\theta) \\
                                 & \qquad + \sum_{i=1}^n \log\,p(y_i|\theta, (L_Q^{-\top}v)_i).
\end{align}
The first two terms of this expression are straightforward to compute. The main computational challenge is the Cholesky decomposition of the precision matrix $Q$ and solving the linear system $f(X)=L_Q^{-\top}v$. We discuss in Section~\ref{sec:op} how these operators (and their derivatives) can be implemented to make HMC efficient for banded precision matrices.\footnote{\citet{faulkner2018locally} report that they successfully used HMC for sampling from GMRF models using the probabilistic programming language Stan~\citep{carpenter2017stan}. Their approach does not take advantage of the sparsity that is found in GMRF precision matrices, so this sampling is not efficient for large models. This specific topic is however currently under discussion between the Stan developers~\citep{simpson2017sparse}.}

\subsection{Variational inference}
\label{subsec:vi}
Variational inference achieves approximate inference by maximising a lower bound to the marginal likelihood over a family of tractable distributions \citep{blei2017variational}. We here derive the necessary banded matrix operations that are needed for {VI}.

Let $F = f(X) \in \RR^n$. We assume a Gaussian prior $F \sim \NN(m_p, Q_p^{-1})$ with banded precision, where $m_p$ and
$Q_p=L_p L_p^T$ may depend on parameters $\theta$.
We also choose a Gaussian distribution $\NN(m_q, Q_q^{-1})$ with banded precision to approximate the posterior $F| Y$.
We denote its probability density by $q$, and we parameterise it by its mean $m_q$ and the Cholesky factor $L_q$ of $Q_q$.

We finally assume that the likelihood factorises to obtain the following log-likelihood lower bound as our variational objective:
\begin{align}
\nonumber    \log\,p(Y) &\geq \EE_{q(F)} \log\frac{p(Y,F)}{q(F)} \\
               & = \sum_{i=1}^n \EE_{q(F_i)} \log \, p(Y_i|F_i) - \KL{q}{p}.
\label{eq:elbo}
\end{align}
The first term of Eq.~\ref{eq:elbo} only depends on the marginal distributions $q(F_i)$, which are described
by $m_q$ and the diagonal values of $Q_q^{-1}$. Although $Q_q^{-1}$ is typically a dense matrix, its diagonal
values can be obtained efficiently with the \textit{sparse inverse subset} method discussed in Section~\ref{subsec:op}.
Depending on the likelihood, $\log \, p(Y_i|F_i)$ may or may not have a closed form. If no analytical
expression is available, the problem typically boils down to the numerical approximation of one-dimensional integrals 
which can be done via Monte Carlo sampling or quadrature methods \citep{hensman2015scalable}.
Regarding the Kullback–Leibler divergence term in Eq.~\ref{eq:elbo}, it can be expressed as:
\begin{align}
    \nonumber    \KL{q}{p} & =\frac{1}{2} \Big( \mathrm{tr} (  Q_q^{-1} Q_p ) + 2 \sum_i \log [L_q]_{ii} - \log [L_p]_{ii}\\
               & \ \  +  (m_p - m_q)^{T} L_p L_p^T  (m_p -m_q)  - \N  \Big). 
\label{eq:KL}
\end{align}
The trace term in this expression can be computed as the sum of an element-wise product between $Q_q^{-1}$ and $Q_p$.
Since $Q_p$ is banded, it is sufficient to compute only the elements of $Q_q^{-1}$ that lie inside the band of $Q_p$.
Here again, we can use the \textit{sparse inverse subset} operator.

%% file: ops.tex
\section{Banded low level operators}
\label{sec:op}
In the previous section we collected a list of the operators needed to perform efficient inference for GP models with banded precisions. 
We now show that these operators can be implemented efficiently with a complexity at most ${\cal O}(\N l^2)$ for 
$\N \times \N$ matrices of bandwidth $l$.
Furthermore, we derive expressions for the reverse mode differentiation of these operators that also have linear complexity in $N$.

\subsection{Description of operators}
\label{subsec:op}

\paragraph{Cholesky decomposition.} $\opC:\ Q \rightarrow L \ \textsl{s.t.} \ LL^T = Q$.
One fundamental property of the Cholesky decomposition of a banded matrix $B$ (assumed symmetric and positive-definite)
is that it returns a lower triangular matrix with the same number of sub-diagonals as $B$. Its implementation for a
banded matrix is similar to the dense case (see Algorithm~\ref{alg:chol} in Appendix~\ref{sec:appendix:algo}) but
its complexity is $\mathcal{O}(\N l^2)$ instead of $\mathcal{O}(\N^3)$, as detailed by \citet{rue2005gaussian}. 

\paragraph{Triangular solve.} $\opS: \ (L, v) \rightarrow L^{-1} v$.
It turns out that the algorithm is similar to a classic triangular solve algorithm running in nested loops through all rows and columns of $L$, but the inner loops on columns can be started at the beginning of the band. This leads to a $\mathcal{O}(\N l)$ complexity instead of $O(\N^2)$ for dense matrices~\cite[p.45]{rue2005gaussian}.

\paragraph{Sparse inverse subset.} $\opI: \ L \rightarrow (Q^{-1})$.
Although the inverse of a banded matrix is often a dense matrix, Takahashi's algorithm~\citep{takahashi1973formation}
shows that it is possible to compute only the band elements of $Q^{-1}$. The pseudo code for this operator is given
by Algorithm~\ref{alg:inv} in Appendix~\ref{sec:appendix:algo}, and it results in a $\mathcal{O}(\N l^2)$ complexity. 

\paragraph{Products.}
The matrix product is another operation that preserves bandedness: the resulting lower bandwidth is the
sum of the lower bandwidths of the inputs (and similarly for the upper bandwidth). This operator is denoted by $\PP: B_1, B_2 \rightarrow B_1  B_2$
and its complexity is $\mathcal{O} (N l^2)$.
We additionally need the following basic linear algebra operations:
\emph{product} between a banded matrix and a vector $\PP: B, v \rightarrow B  v$, that is $\mathcal{O}(Nl)$;
and \emph{outer product} of two vectors ($\OO : m, v\rightarrow mv^T$). The latter typically yields a dense matrix and it
has a $\mathcal{O}(N^2)$ complexity. Although this may seem problematic, we only require in our applications a small band
of this dense matrix, which can be computed with a cost that is linear in $N$.

Note, also, that although we are primarily interested in lower-banded matrices, various operations require some matrix \emph{transposes}
(examples below with the expressions for various gradients).  
This forces the implementation to deal with several variants of each algorithm, 
such as solving linear systems with lower-banded or upper-banded matrices.

\subsection{Derivatives of the operators}
\label{subsec:derivatives}

We endow each operator with a method implementing its \emph{reverse-mode differentiation} (see Appendix~\ref{sec:appendix:rev_mode}). Given a chain of operations resulting in a scalar value (say $X \to Y \to c$), it consists in propagating a downstream gradient ($\frac{dc}{dY}$) to an upstream gradient ($\frac{dc}{dX}$). 
This approach has two main advantages: 
(1) it is \emph{compositional}, which allows the gradients to be obtained for arbitrarily complex models based on our banded matrix operators; 
(2) it is \emph{efficient}: the execution time of the reverse mode differentiation of a model takes a time proportional to its 
forward evaluation.

Following the literature \citep{giles2008collected,murray2016differentiation}, we denote by $\bar{X}$ 
the ``reverse-mode sensitivities'', or gradients computed in reverse-mode on the output $X$ of an operator. In our previous example, $\bar{X}=\frac{dc}{dX}$.

\paragraph{Basic operators}

The expressions of the gradients of all product and solve operators are derived in Appendix~\ref{sec:appendix:rev_mode}
and summarised in Table~\ref{tab:derivatives}. They can all be defined as a simple composition of the forward evaluation
of banded operators. For example, the gradient of a \emph{solve} is defined using solve and product operations only. 

\begin{table*}
\caption{Summary of the reverse mode sensitivities with analytical expression.}
\begin{center}
{\renewcommand{\arraystretch}{1.2}
\begin{tabular}{ccccc } 
\hline 
Operator & Symbol & Input & Forward &  Reverse Mode Sensitivities  \\ 
\hline
product matrix-matrix & $\opP$ 
& $B_1$, $B_2$ & $P = B_1  B_2$ 
& \spc{$\bar{B_1} = \opP(\bar{P}, B_2^{T})$
       \qquad $\bar{B_2}=\opP(B_1^{T}, \bar{P} )$ }
\\ 
product matrix-vector & $\opP$ 
& $B$, $v$ & $p = B v$ 
& \spc{$\bar{B}= \opO(\bar{p}, v)$  \quad $\bar{v}=\opP(B^{T}, \bar{p} )$ }  
\\ 
vector outer product & 
 $\opO$ & $m, v $ &  $O = mv^T$ 
 & \spc{$\bar{m} = \opP(\bar{O}, v)$ \quad $\bar v = \opP(\bar{O}^T, m)$ }
\\
solve matrix-vector & $\opS$ & $L$, $v$ & $ s = L^{-1}  v$ &  \spc{%
    $\bar{v} =\Ss( L^{T},\bar{s})$ 
    \\
	$\bar{L}^{T} =  -\opO(\opS(L, v), \opS(L^T, \bar{s}))$
    }
\\
solve matrix-matrix & $\opS$ & $L$, $B$ & $ S = L^{-1} B$ &  \spc{%
    $\bar{B} =\opS( L^{T },\bar{S} ) $  
    \\
	$\bar{L}^{T} =  -\opP(\opS(L, B), \opS(L^T, \bar{S})^T)$
    }
\\ 
 \hline 
\end{tabular}}
\end{center}
\label{tab:derivatives}
\end{table*}
A few points are worth noting: 
(1) The notation introduced in Table~\ref{tab:derivatives} is overloaded: $\PP$ denotes, for instance, a banded product where the right-hand side 
denotes either a banded matrix or a vector, which is clear from the context.
(2) All intermediate terms in the expressions of the reverse-mode sensitivities 
column can be kept banded, throughout their evaluation. Consequently, the gradients of all operations can be computed in time
proportional to the forward evaluation, which means that the gradients in Table~\ref{tab:derivatives} have at most a ${\cal O}(\N l^2)$ complexity.

\paragraph{Cholesky and sparse subset inverse.}

To define the gradients of the Cholesky and subset inverse operators, we had to use a different approach. 
For both operators, one can define an analytical expression for the reverse-mode sensitivities
($-2\Ss(L, \PP(\bar{\Sigma}^T, (LL^T)^{-1}))$ for subset inverse; see \citet{murray2016differentiation} for Cholesky).
However, evaluating these terms requires the computation of dense matrices and scales as ${\cal O}(N^3)$. Such an approach would thus
lead to a computational bottleneck and render our scheme less efficient.

For the Cholesky reverse-mode differentiation we used an existing gradient computation algorithm that was easy to adapt 
to a banded representation \citep{giles2008collected}, and is described as Algorithm~\ref{alg:chol_bar} in our Appendix.
For the sparse subset inverse operator, we did not find a gradient computation algorithm in the literature that could be customized
for banded matrices. We therefore used the Tangent software package~\citep{van2018tangent}, to generate
a gradient computation algorithm from the forward computation code. We then hand-curated the generated code, obtaining 
Algorithm~\ref{alg:inv_bar} in Appendix~\ref{sec:appendix:algo}.
Both algorithms have a ${\cal O}(\N l^2)$ complexity.

\subsection{Storage footprint}
Using Gaussian process models usually requires the storage of covariance matrices of size $\n \times \n$, where $\n$ is the number of observations. This is usually the limiting factor, and the maximum number of observations that can be handled on currently available desktop computers is typically in the range $\n \in [10^4, 10^5]$.

When working with symmetric or lower triangular banded matrices, it is of course sufficient to store only the non-zero elements of the lower half of the matrix. In our implementation of the operators described above, all the operators use the following convention for their inputs and outputs: let $B$ be a banded matrix of size $n \times n$ which has lower and upper bandwidths equal to $(l_l, l_u)$, we store $B$ as an $(l_l + l_u + 1) \times n$ matrix $B'$ with $B_{i, j} =  B'_{i-l_u-j+1,j}$ for $1 \leq i, j \leq n$. Note that the values of $B'$  located in the upper-left and lower-right corners may not be defined but they are never accessed by algorithms in practice.

%% file: experiments.tex
\section{Experiments}
\label{sec:experiments}

\subsection{Implementation}

We have implemented the banded operators described in Section \ref{sec:op} as custom operators for TensorFlow.
This allowed us to experiment with complex GP models using these operators together with the 
functionalities of the GPflow library~\citep{matthews2017gpflow}.

Our implementation is based on TensorFlow's extensibility mechanism referred to as ``custom ops'': 
the code for the forward evaluation of each operator is written in C++ and registered to TensorFlow,
together with a mix of Python and C++ code that implements the gradients of each operator.
The C++ code for forward evaluation is a direct implementation of the algorithms detailed in Appendix \ref{sec:appendix:algo}.
Most gradients can be written in Python by calling other operators, following Table \ref{tab:derivatives}.
The Cholesky and sparse subset operators require dedicated C++ code, 
as explained in Section \ref{subsec:derivatives} and detailed in Appendix \ref{sec:appendix:algo}.

\subsection{Computational time}
\label{subsec:kalman}

The aim of this section is to confirm that models written using our operators are faster to train than existing alternatives. 
We have seen previously that the complexity of
the proposed operators is $\mathcal{O} (Nl^2)$ where $N$ is the size of the precision and $l$ is the bandwidth.
We now illustrate the influence of these parameters on the time required for computing
the log-likelihood and its gradient for a GP regression model with Gaussian likelihood.

The weekly average atmospheric CO$_2$ concentrations recorded at the Mauna Loa Observatory,  Hawaii, by \citet{keeling2005atmospheric} are commonly
used to demonstrate how different patterns observed in the data (e.g. periodicity, increasing trend)
can be encoded in a GP prior by designing compositional kernel functions~\citep{rasmussen2006gaussian}. We ensure in this example the bandedness of the precision matrix by working within the class of kernels that have a state-space representation~\citep{grigorievskiy2016parallelizable}. %
Conveniently, finite sums and products of such kernels belong to this class (with each composition increasing the resulting state-space dimension). For this experiment, we design our kernel as follows: to capture the slow varying trend we use a Mat\'ern\sfrac{3}{2} kernel $k_{s}(\tau)$ parameterized by a lengthscale $l_s$ and a variance $\sigma_s^2$; for the quasi-periodic trend, we follow \citet{solin2014explicit} and use the quasi-periodic kernel $k_{q\text{-}per}(\tau) = k_q(\tau)\sum_{j=1}^J \cos(2 \pi j f_0 \tau )$, where $f_0$ is the frequency of the periodic trend and $k_q$ is a Mat\'ern\sfrac{1}{2} kernel with lengthscale $l_q$ and variance $\sigma_q^2$. These kernels have state-space dimensions of respectively $2$ and $2J$, so the state dimension of $k = k_s + k_{q-per}$ is $2J + 2$ and the bandwith of the resulting precision matrix is $4 J + 3$.

The regression model using this kernel can be implemented in three different ways that we are going to compare: the first one is a Kalman filter implementation using loops in TensorFlow; the second one is based on our custom operators and uses the bandedness of the model's precision matrix; and the third one is the GPflow implementation of classic GP regression which uses the dense covariance matrix. Note that the first two implementations exploit the Markov property of the model and are thus expected to scale linearly with the amount of data points, whereas the third is known to scale cubically. These algorithms are implemented using TensorFlow and are detailed in Appendix~\ref{sec:appendix:pseudocode_algos}.

\begin{figure*}[!htb]
\minipage{0.32\textwidth}%
  \includegraphics[width=\linewidth]{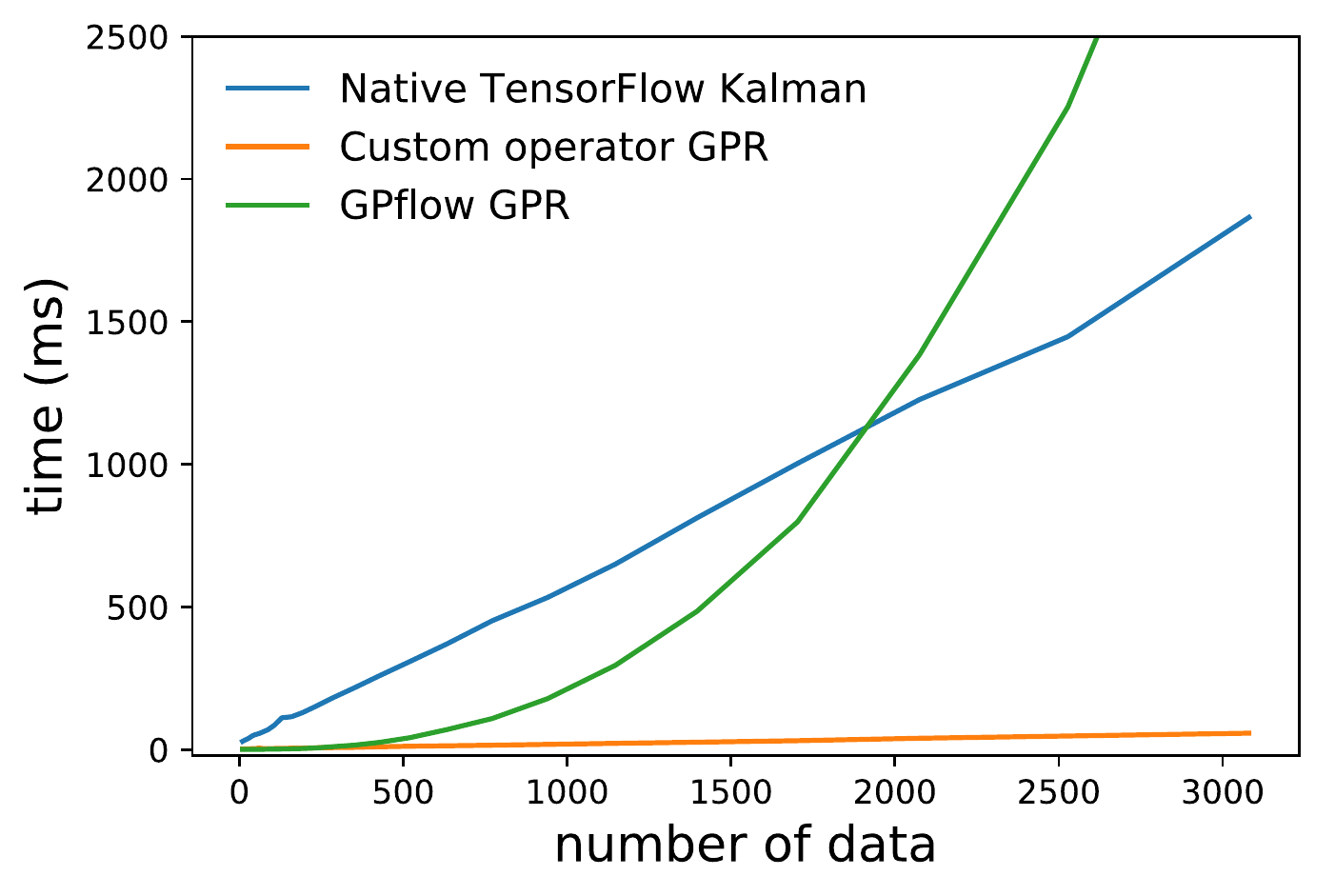}
\endminipage
\minipage{0.32\textwidth}
  \includegraphics[width=\linewidth]{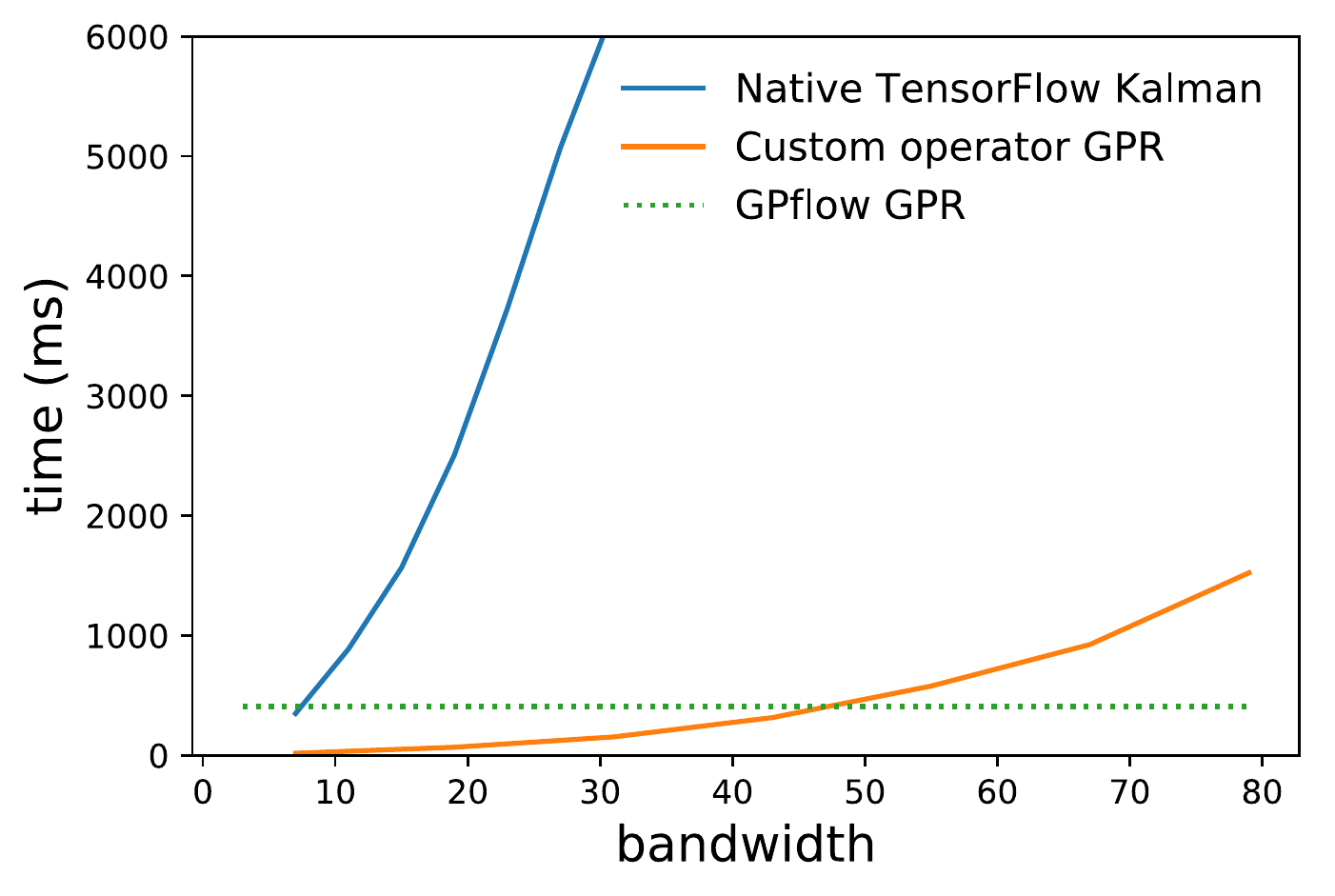}
\endminipage\hfill
\minipage{0.32\textwidth}
\includegraphics[width=\linewidth]{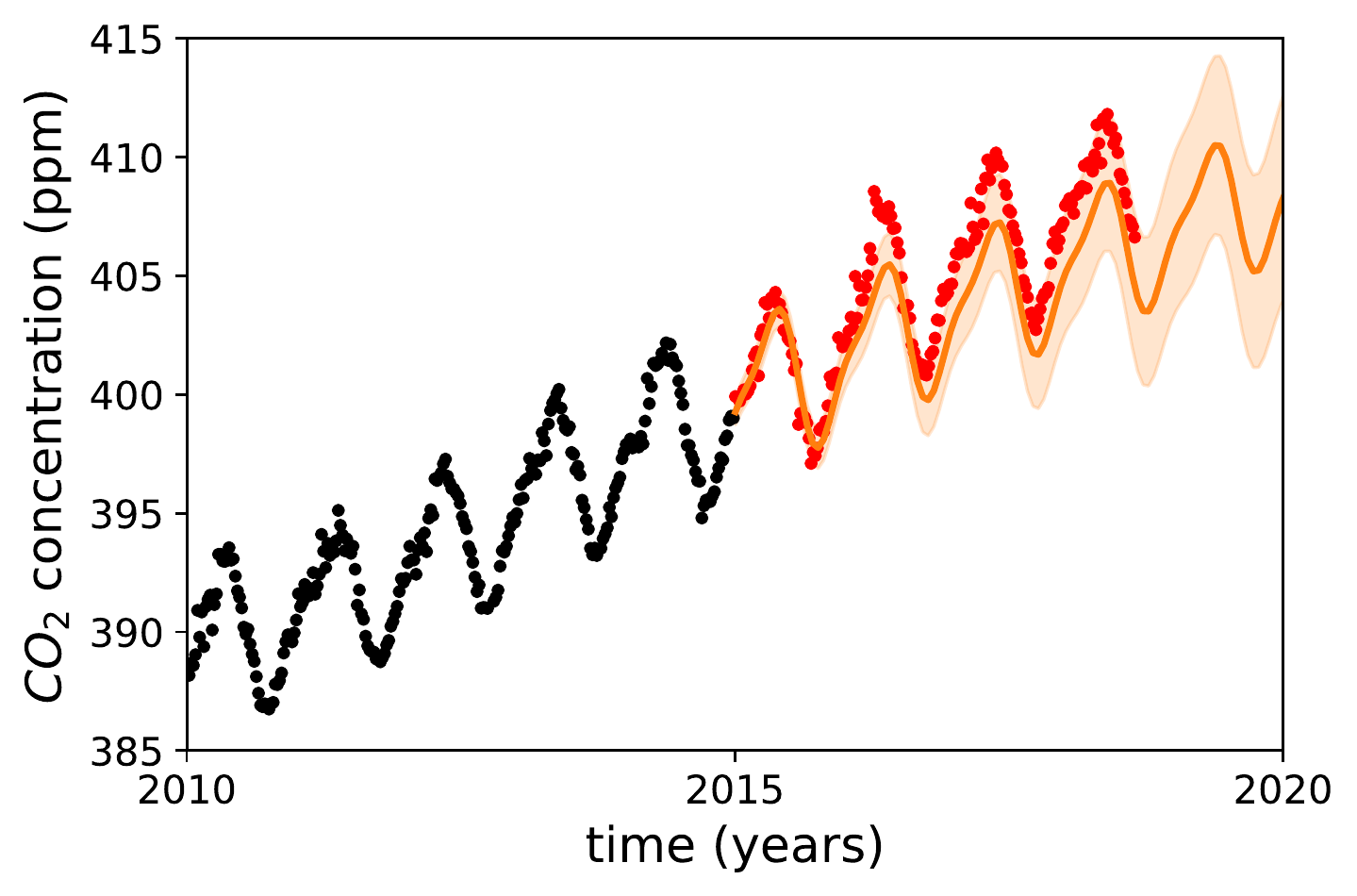}
\endminipage\hfill

\caption{
Mean execution time for three implementations of a GP regression model as a function of the number of data points (left) and of the bandwidth (middle). The right panel shows the model predictions when the data points after 2015 are excluded from the training dataset.  
}
    \label{fig:timing}
\end{figure*}

The dataset contains $n=3082$ observations and we consider subsets of increasing sizes to time the evaluation of the marginal likelihood and its gradient with respect to the model parameters (Figure~\ref{fig:timing}, left). As expected, the computational time for the naive GP regression models quickly becomes prohibitively large whereas the implementations based on the state-space representation are linear in time. The computational speed-up for custom operators is of three orders of magnitude for the full dataset. 

Similarly, we show in the middle panel the influence of the bandwidth $l$ (which is twice the state-space dimension $d$ minus one) on the execution time, restricting the dataset to the first $n=1500$ points. To do so we vary the state dimension of the model by increasing $J$, that is by adding more harmonics to our quasi-periodic kernel. One can see that our banded operators are much faster than the Kalman implementation, and that it favourably compares to the GPflow implementation when the bandwidth is smaller than 45. Furthermore, this threshold would increase quickly as the dataset gets larger. With our operators, the likelihood evaluation scales as $\mathcal{O}(Nl^2)$, but we now have $N = nd$ so the operators complexity is $\mathcal{O}(nl^3)$. Note that this is the same complexity as the classic implementation of the Kalman filter.

Finally, Figure~\ref{fig:timing} (right) shows the model predictions ($J=2$, $l=11$) for the years from 2015 to 2020. We use the implementation based on our custom operators and we learn the kernel parameters by optimizing the marginal likelihood of the model given the data from 1958 until 2015. This illustrates that the model can account for complex patterns even with a small bandwidth. The mean test log-likelihood of this model is -1.75 whereas we obtain -1.56 with the reference implementation \citep[Eq. 5.19]{rasmussen2006gaussian}. Although, this is slightly to the advantage of the later, it means that a model with small bandwith can have good prediction abilities, even when it is not finely tuned for the dataset at hand.

\subsection{Gaussian Markov random field}
In this section we illustrate our ability to perform inference on a GMRF with non-conjugate likelihood.
To this aim, we consider the Porto dataset that gathers the GPS locations of taxi pick-ups in the city of Porto for the period July 2013 - June 2014. We use the first three weeks of the data as our training set and the following three weeks as our test set. This dataset has already been modeled successfully with GP based Cox processes by~\cite{john2018large} but we choose a different approach here: we consider a GP based Cox process model defined on a graph representing the road network and each data point is projected onto the closest node (if it is within a 10m radius). The main advantage of this approach is that the GP covariances are using the graph distance, which are more meaningful that the Euclidian distance (think about two locations separated by the river). 

The graph is an undirected graph obtained from open street map, it is denoted by $G = (V = \{1, \dots, \N\}, E)$ and it consists of $\N = 11284$ nodes and $\#V=12185$ edges. The length (in meters) of the edge $(i, j) \in E$ is denoted by $d_{i,j}$.

Let $f \sim \NN (0, Q^{-1})$ be a latent GMRF indexed by the nodes $v \in V$. Our generative model assumes that the number of pick-up associated to a node $i$ follows a Poisson distribution with parameter $\exp(f_i)w_i$, where $w_i$ is the length of the edges associated to the node $i$: $w_i = \sum_{j, (i, j) \in E} \max(10, d_{i,j}/2)$. 

Since $Q$ can also be interpreted as an inner product for vectors $g, h \in \mathds{R}^\N$: $\langle g, h \rangle = g^T Q h$, we define $Q$ such that it corresponds to the sum of Mat\'ern\sfrac{1}{2} inner products over all the edges~\citep{durrande2016detecting}:
\begin{align}
\nonumber    g^T Q h &= \frac{1}{\sigma^2} \sum_{(i, j) \in E} \frac{1}{1 - \lambda_{i,j}}(g_i \ g_j) 
    \begin{pmatrix}
        1 & -\lambda_{i,j}\\
        -\lambda_{i,j} & 1
    \end{pmatrix}
    \begin{pmatrix}
        h_i\\
        h_j 
    \end{pmatrix}\\
    & \qquad \qquad \qquad -\frac12 g_i h_i - \frac12 g_j h_j
\end{align}
where $\lambda_{i,j} = \sigma^2 \exp (-d_{i,j} / \ell)$ with $\sigma^2=10$, $\ell=10^4$.

We now compare three methods for predicting the values of the latent function $f$ given the observations of $y_v$ for  $v \in V$: a Hamiltonian Monte-Carlo sampler, a variational inference method, and a baseline consisting in estimating the rate of a Poisson random variable independently for each node. 
In the first two cases, we use our implementation based on the GPflow framework together with the specialised operators for banded matrices described in Section \ref{sec:op}. 
The first step before actually building the models consists in finding a good ordering of the nodes in order to reduce the bandwidth of the precision matrix. 
Using the Cuthill McKee algorithm from the Scipy library, we found an ordering corresponding to a lower bandwidth of $l=117$ for matrix $Q$.

Since our inference methods has a runtime of ${\cal O}(\N l^2)$, as opposed to ${\cal O}(\N^3)$ for a dense representation,
the settings $\N=11284$ and $l=117$ imply that our banded framework saves runtime by a factor of roughly $10^4$. Similarly, having to store only the band of the matrices instead of their dense versions allows us to save almost two orders of magnitude on the storage footprint.

\begin{figure}[h!]
    \centering
    \begin{subfigure}[b]{0.49\textwidth}
        \includegraphics[width=\textwidth]{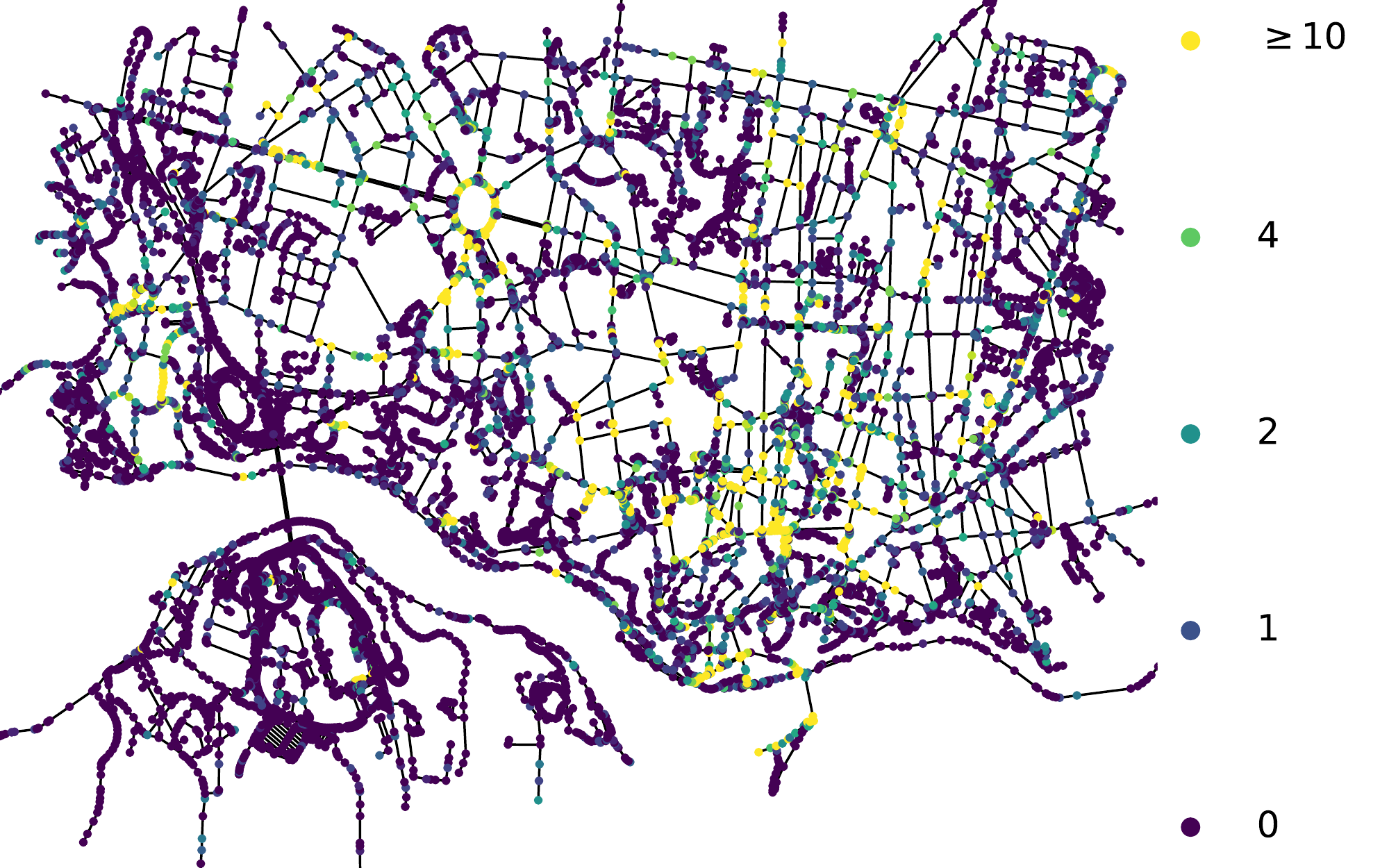}
        \caption{Data: number of pick-ups per node.\\}
        \label{fig:porto_data}
    \end{subfigure}
    \\ \ \\
    \begin{subfigure}[b]{0.49\textwidth}
        \includegraphics[width=\textwidth]{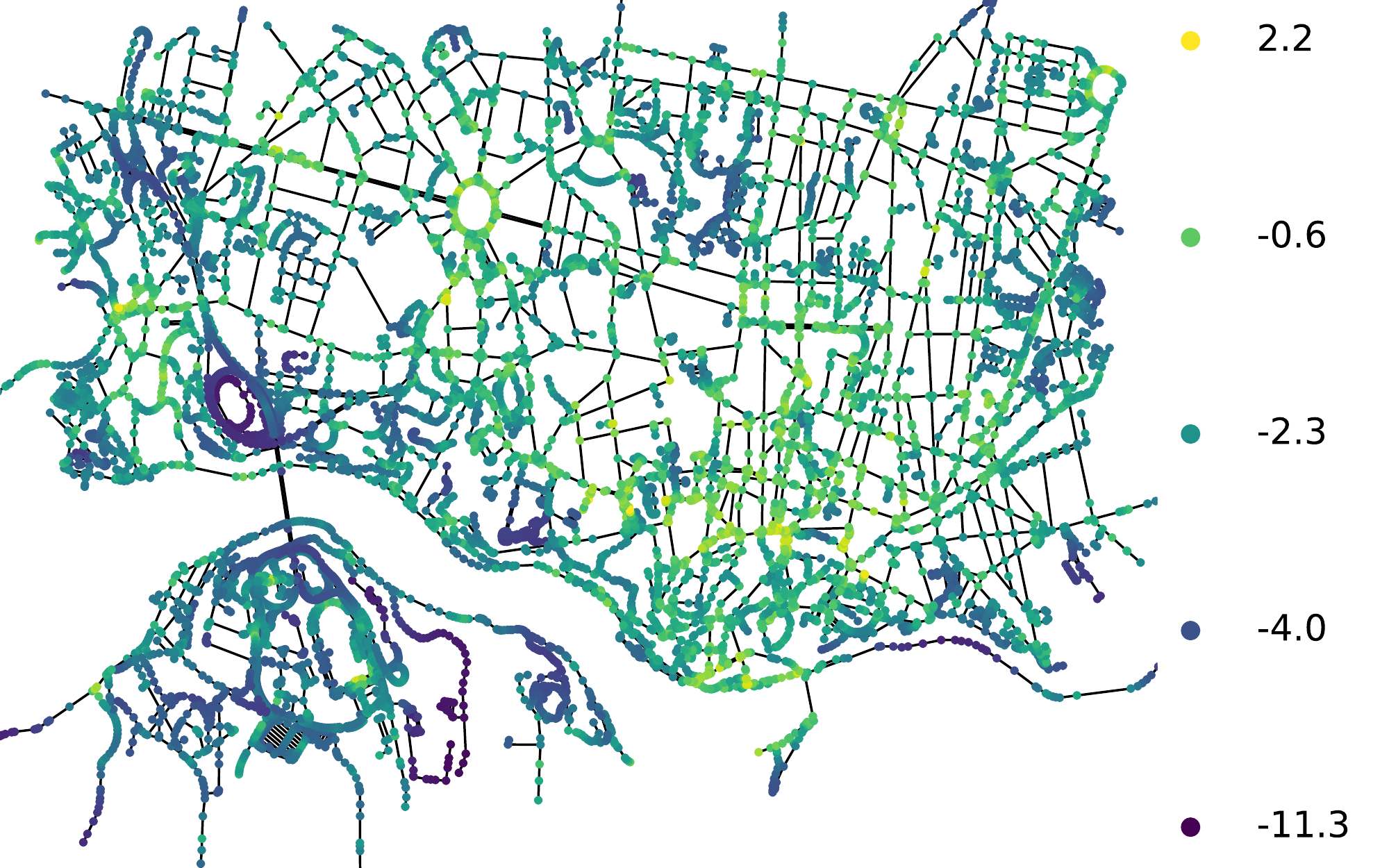}
        \caption{Predicted latent.\\}
        \label{fig:HMC_mean}
    \end{subfigure}
    \caption{Graphs of the Porto experiment. (a) Number of pick-ups after clipping them to the graph. The range of the data is [0, 160] but we choose a non-linear color map. (b) Predicted latent function for the VI model.}
    \label{fig:porto}
\end{figure}

Figure~\ref{fig:porto} (bottom) shows the mean prediction of the model trained using variational inference. One can see that the model successfully extracts a smooth trend for the latent variable $f$. However the non-linear mapping from the latent function to the rate $\lambda_i = \exp(f_i)w_i$ leads to a large range of predicted rates: for most nodes, the predicted rate is below 5, but its maximum value is 149. We investigated some of the locations with large predicted rate and they all correspond to particular landmarks such as hotels or hospitals where the taxi demand is naturally high and sharp.
The plots we obtain for the HMC and the baseline predictions are very similar so we do not reproduce them here.

Finally, we compare the likelihood of the three models on the test set. The values of the log-likelihood for the VI, HMC and the baseline are -15778.5, -15873.6 and -17146.6 respectively. This shows that even for this challenging dataset, the proposed model has powerful predictive power.

\input{conclusion.tex}

\bigskip

%% file: conclusion.tex
\section{Discussion and conclusion}

This work has examined how some Gaussian Markov models can be expressed using banded
matrices. We investigated some inference and end-to-end learning procedures for these models and
identified the shared set of general banded operators---endowed with their reverse-mode derivatives---necessary to implement them. 

The framework we propose is general in the sense that it covers a large class
of models (Kalman, SSM, GMRF) for which it provides several state of the art
inference methods such as variational inference, gradient-based samplers
and maximum likelihood estimation when a subset of the variables are observed.
More inference algorithms such as expectation propagation readily fit into this paradigm.

The only algorithm we could not find in the literature is the differentiation
of the sparse inverse subset $\opI$. Given that all others were readily
available, we believe that the implementation of a few low level operators is a
small price to pay for the huge benefit they provide in practice.
Although we focused on matrices where the precisions are banded, similar
work could be carried out for matrices with other sparsity patterns such as the
ones obtained when the nested dissection is used for the graph node
ordering~\citep{rue2005gaussian}.

We anticipate two main outcomes for the present work. The first one is a
strong incentive for a better support for the sparse algebra in
automatic differentiation frameworks, and the second is a renewal of the popularity
of Gaussian Markov models now that state of the art inference methods are
available.

%% file: appendix_derivations.tex
\section{Reverse mode sensitivities derivations}
\label{sec:appendix:rev_mode}
We are interested in reverse mode differentiation. That is, we propagate the gradient backward in the chain of operations from the scalar objective to the upstream parameters.

For a chain of operations $X \to Y \to \dots \to c$ (with $X,Y$ matrices), we have 
\[dc = \sum_{ij} \frac{\partial c}{\partial Y_{ij}} dY_{ij}\]
Noting $\bar{Y} = \left[\frac{\partial c}{\partial Y_{ij}}\right]$, we can rewrite this as $dc = Tr(\bar{Y}^T dY)$. When doing reverse mode differentiation, we propagate the errors from the objective. 
From the mapping from $X \to Y$, we figure out matrices $L_X$ and $R_X$ such that $dY = L_X(dX)R_X$. It follows that $dc = Tr(\bar{Y}^T \, L_X(dX)R_X)$ and, after a rotation in the trace, we can then identify $\bar{X}^T = R_X\bar{Y}^TL_X$.

\subsection{Outer product $m, v \to m v^T$}

$o = m v^T$, so $\mathrm{d}o = \mathrm d m\, v^T + m\, \mathrm d v^T$ \\
For fixed $m$: $\mathrm d o = m\, \mathrm d v^T$ so 
\begin{align*}
\bar{v}  = \bar{o}^T m.
\end{align*}
For fixed $v$: $\mathrm d o = \mathrm d m v^T$ so 
\begin{align*}
\bar{m}^T &= v^T \bar{o}^T, \\
\bar{m} &= \bar{o} v.   
\end{align*}
 
\subsection{Product $ B_1^X, B_2^Y \to B_1^X B_2^Y$}

$X$ and $Y$ are both either $T$ (for transpose) or $1$ (for identity).
This allows to write all cases of argument transposition at once.\\
$ P = B_1^X B_2^Y$, so $\mathrm d P = \mathrm d (B_1^X)B_2^Y + B_1^X \mathrm d (B_2^Y) $.\\
For fixed $B_1^X$: $\mathrm d P = B_1^X \mathrm d (B_2^Y) $ so 
\begin{align*}
    \bar{B}_2^{YT} &= \bar{P}^T  B_1^X = \opP( \bar{P}^T , B_1^X ). 
\end{align*}
For fixed $B_2^Y$: $\mathrm d P = \mathrm d (B_1^X) B_2^Y $ so 
\begin{align*}
    \bar{B}_1^{XT} &=  B_2^Y   \bar{P}^T =  \opP( B_2^Y,   \bar{P}^T).
\end{align*}

\subsection{Solve $L^X, B^Y \to L^{-X} B^Y$}

$X$ and $Y$ are as defined in the product section.\\

$ S = L^{-X} B^Y$, so 
\begin{align*}
\mathrm d S &= \mathrm d (L^{-X})B^Y + L^{-X}\mathrm d (B^Y) \\
&= -L^{-X}\mathrm d (L^X)L^{-X}B^Y + L^{-X}\mathrm d (B^Y). 
\end{align*}
For fixed $B^Y$: $\mathrm d S = -L^{-X}\mathrm d (L^X)L^{-X}B^Y$ so $\bar{L}^{XT} = -L^{-X}B^Y \bar{S}^T L^{-X}$ \\
\begin{align*}
\bar{L}^{XT} &= -L^{-X}B^Y \bar{S}^T L^{-X}\\
	&= -(L^{-X}B^Y) (L^{-XT} \bar{S})^T \\
	&= -\opP(\opS(L^X, B^Y), \opS(L^{XT}, \bar{S})^T).
\end{align*}
For fixed $L^X$: $\mathrm d S =  L^{-X}\mathrm d (B^Y)$ so 
\begin{align*}
\bar{B^Y}^T &=\bar{S}^T L^{-X}\\
\bar{B^Y} &= L^{-XT }\bar{S} \\
      &= \Ss( L^{XT },\bar{S} ).
\end{align*}

\subsection{Product $B,v  \to B v $}

$p = Bv$, so	$\mathrm d p = \mathrm d B v + B \mathrm d v $.

For fixed $v$, $\mathrm d p = \mathrm d B v$, so
\begin{align*}
\bar{B}^T &=v \bar{p}^T \\
\bar{B} &=\opO(\bar{p}, v^T). 
\end{align*}

For fixed $B$, $\mathrm d p = B \mathrm d v$, so
\begin{align*}
\bar{v}^T &= \bar{p}^T B \\
\bar{v} &= \opP(B^T, \bar{p}).  
\end{align*}

\subsection{Solve $L,v  \to L^{-1} v $}

$u = L^{-1}v$, so
\begin{align*}
	\mathrm d u &= \mathrm d (L^{-1}v) \\
	&= \mathrm d (L^{-1})v + L^{-1} \mathrm d v \\
	&= -L^{-1} (\mathrm d L) L^{-1}v + L^{-1} \mathrm d v.
\end{align*}

For fixed $L$, $\mathrm d u = L^{-1} \mathrm d v $, so
\begin{align*}
\bar{v}^T &= \bar{u}^T L^{-1} \\
&= \opS(L^T, u)^T.
\end{align*}
For fixed $v$, $\mathrm d u = -L^{-1}(\mathrm d L)L^{-1}v $, so
\begin{align*}
\bar{L}^T &= - L^{-1}v \bar{u}^T L^{-1}\\
& = \opP( \opS(L, v), \opS(L^T, u)^T ).
\end{align*}

%% file: appendix.tex
\section{Algorithms}
\label{sec:appendix:algo}
We report algorithms operating on a \emph{dense} representation of the banded matrices (the usual $i,j$ indexing). In practice these only read and write into the band of the matrices involved as input or output. A derivation using the (\emph{diag, column}) indexing would reflect this fact more clearly but blurs the inner working of the algorithms.   
In a similar spirit, our algorithms sometimes include avoidable instantiations of temporary variables that are kept for clarity.
For reverse mode differentiation we refer the reader to \citep{giles2008collected} from which we follow the notation.

\newcommand{\keyword}[1]{}
\renewcommand{\keyword}[1]{~\textbf{\emph{#1}}~}

\begin{algorithm}[H]
 \SetKwInOut{Input}{Input}  
 \SetKwInOut{Result}{Output}  
 \Input{\ $Q$: an $n \times n$ symmetric positive definite matrix with $l$ lower sub-diagonals.}
 \Result{\ $L$: an $n \times n$ lower triangular matrix with $l$ lower sub-diagonals.}
    \vspace{2mm}
    Initialize L with zeros \\
    \For{i \keyword{in} $0 \keyword{to} n-1$}{
        \For{j \keyword{in} $\max(i-l,0) \keyword{to} i+1$}{
            $m$ = $\max(i-l, j-l, 0)$ \\
            $s$ = sum$(L[i,m\!:\!j] * L[j,m\!:\!j])$ \\
            \eIf{i==j}{
                $L[i,j]$ = sqrt($Q[i,j] - s$) \\
            }{
                $L[i,j]$ = $(Q[i,j] - s) / L[j,j]$\\
            }
        }
   }
 \caption{Cholesky algorithm for banded matrices.}
 \label{alg:chol}
\end{algorithm}

\begin{algorithm}[H]
 \SetKwInOut{Input}{Input}  
 \SetKwInOut{Result}{Output}  
 \Input{\ $L$: the $n \times n\,$ Cholesky factor of $Q$, with $l$ lower sub-diagonals. \\
    \ $L\_b$: the $n \times n$ reverse mode derivatives of $L$ with respect to the objective function.}
 \Result{\ $Q\_b$: the $n \times n$ reverse mode derivatives of $Q$ with respect to the objective function.}
    \vspace{2mm}
    Initialize $Q\_b$ with zeros \\
    \For{i \keyword{in} n-1 \keyword{downto} 0}{ 
        $j\_stop = \max(i-l, 0)$\\
        \For{j \keyword{in} i \keyword{to} j\_stop}{
            \eIf{j == i}{ 
                $Q\_b[i, i] = 1/2 * L\_b[i, i] / L[i, i]$
            }{
                $Q\_b[i, j] = L\_b[i, j] / L[j, j]$ \\
                $L\_b[j, j] -= L\_b[i, j] * L[i, j] / L[j, j]$
            }
            \For{l \keyword{in} j-1, \keyword{in} j\_stop}{
                $L\_b[i, l] -= Q\_b[i, j] * L[j, l]$\\
                $L\_b[j, l] -= Q\_b[i, j] * L[i, l]$
            }
       }
    }
 \caption{Reverse mode differentiation for the Cholesky operator for banded matrices.}
 \label{alg:chol_bar}
\end{algorithm}

\begin{algorithm}[H]
 \SetKwInOut{Input}{Input}  
 \SetKwInOut{Result}{Output}  
 \Input{\ $L$: an $n \times n$ lower triangular matrix with $l$ lower sub-diagonals. }
 \Result{\ $S$: an $n \times n$ lower triangular matrix with $l$ lower sub-diagonals of $[L L^T]^{-1}$.}
    \vspace{2mm}
    Initialize \textsl{Sym} as a (symmetric) $n \times n$ banded matrix of \emph{both} lower and upper bandwidth $l$ \\
	$\textsl{vec} = \textrm{diag}(L)$ \\
	$U =  \textrm{transpose}(L / \textsl{vec})$ \\
    \For{j \keyword{in} $n-1, \keyword{downto} 0$}{
        \For{i \keyword{in} $j \keyword{downto} max(j - l + 1, 0)$}{
			$\textsl{Sym}[i, j] = -\textrm{sum}(U[i, i + 1:i + l]  \textsl{Sym}[i + 1:i + l, j])$ \\
			$\textsl{Sym} [j, i] = \textsl{Sym} [i, j]$ \\
            \If{i==j}{
				$\textsl{Sym} [i, i] += 1/(\textsl{vec} [i])^2$ \\
            }
        }
    }
	$S = \textrm{lower\_band}(\textsl{Sym})$
 \caption{Algorithm for the inverse of a matrix from its Cholesky decomposition.}
 \label{alg:inv}

\end{algorithm}

\newpage

\begin{algorithm}[H]
 \SetKwInOut{Input}{Input}  
 \SetKwInOut{Result}{Output}  
 \Input{\ 
    $L$: $L$: the $n \times n$ Cholesky factor of $Q$, with $l$ lower sub-diagonals. \\
    \ $S$: The $n \times n$ output of the subset inverse from Cholesky with $l$ lower sub-diagonals. \\
    \ $bS$: the $n \times n$ reverse mode derivatives of $B$ with respect to the objective function. \\
    \emph{Note that both S and bS should be treated as \emph{symmetric} banded matrices, \\
    with bS copied locally (modified in place).}
    }
 \Result{\ $bL$: $n \times n$ the reverse mode derivatives of $L$ with respect to the objective function.}

    \vspace{2mm}
    Initialize $bU$ as a banded matrix similar to $U$, filled with zeros. \\
	Initialize $\textsl{bvec\_inv\_2}$ as a vector of size $n$ filled with zeros. \\
    ~ \\
	$\textsl{vec} = \textrm{diag}(L) $\\
	$U = \textrm{transpose} (L / \textsl{vec}) $\\
    ~ \\
    \For{j \keyword{in} $0 \keyword{to} n-1$}{
        \For{i \keyword{in} $\max(0, j-k+1) \keyword{to} j$}{
            \If{i == j}{
				$\textsl{bvec\_inv\_2}[i] += bS[i, i]$
            }
            \textsl{\% Grad of: $S[j, i] = S[i, j]$} \\
			$\textsl{tmp} = bS[j, i]$ \\
            $bS[j, i] = 0$ \\
			$bS[i, j] += \textsl{tmp}$ \\
            ~ \\
			\textsl{\% Grad of: $S[i, j] = -\textrm{sum}(U[i, i+1:i+k] S[i+1:i+k, j])$} \\
            $bU[i, i+1 : i+k] -= S[i+1:i+k, j] * bS[i, j]$ \\
            $bS[i+1:i+k, j] -= U[i, i+1:i+k] * bS[i, j]$ \\
            $bS[i, j] = 0$
        }
    }
    ~ \\
	\textsl{\% Grad of: $U = \textrm{transpose}(L/\textsl{vec})$} \\
	$\textsl{bL} = \textrm{transpose}(bU) / \textsl{vec}$ \\
    \textsl{\% Grad of $1 / \textsl{vec}^2$} \\
	$\textsl{bvec} = -2 * \textsl{bvec\_inv\_2} / \textsl{vec}^3$ \\
	\textsl{\% Grad of: $1 / \textsl{vec}$} \\
	$\textsl{bvec} -= \textrm{sum}(\textrm{transpose}(bU) L, 0) / (vec^ 2)$ \\
	\textsl{\% Grad of: $\textsl{vec} = \textrm{diag}(L)$} \\
	$bL += \textrm{diag}(\textsl{bvec})$ \\

 \caption{Reverse mode differentiation of the inverse from Cholesky.}
 \label{alg:inv_bar}

\end{algorithm}

\begin{algorithm}[H]

 \SetKwInOut{Input}{Input}  
 \SetKwInOut{Result}{Output}  
 \Input{\    
    $L$: An $n \times n$, lower-triangular banded matrix with $l$ sub-diagonal\\ 
    \ $R$: An $n \times n$ banded matrix with $u$ upper-diagonals \\
    }
 \Result{\ $O$: An $n \times n$ with $l$ lower and $u$ upper sub-diagonals of $L^{-1} R$.}

    \For{k \keyword{in} -$u$ \keyword{to} $l$}{
        \For{i \keyword{in} $\min(n + k - 1, n - 1)$ \keyword{downto} $\max(0, k)$}{
			r = \keyword{if} $(i, i-k) \in \textsl{band}(R)$ \keyword{then} $R[i, i-k]$\keyword{else} 0 \\
			$\textsl{dot\_product} = \textrm{dot}(L[i, :], O[:, i-k])$ \\
			$O(i, i - k) = (r - \textsl{dot\_product}) / L(i, i)$
        }
    }
    \vspace{2mm}

 \caption{Solve of a lower triangular banded matrix by an arbitrary banded matrix.}
 \label{alg:solve}

\end{algorithm}

%% file: appendix_algo_pseudocode.tex
\newcommand{\prior}{0}
\newcommand{\lik}{1}
\newcommand{\post}{2}
\newcommand{\eeta}{u}
\newcommand{\T}{\top}

\section{Pseudocode of algorithms}
\label{sec:appendix:pseudocode_algos}

We here provide details on how we used our operators for banded matrices in the algorithms we proposed. Implementation details are also given for the pre-existing algorithms we used as comparisons.

\subsection{log likelihood evaluation (for HMC) }

In section \ref{subsec:hmc}, we explained how to perform gradient based MCMC in GMRF models. Algorithm \ref{algo:hmc} gives some details on how we evaluate the likelihood given sampled parameters $\theta$ and whitened latent $v$.

\begin{algorithm}[h!]
   \caption{Log Likelihood evaluation}
   \label{algo:hmc}
   
\SetKwInOut{Input}{Input}  
\SetKwInOut{Result}{Output}  
\Input{\ white noise sample $v$: vector of length $n$\\
\ observations $Y$: vector of length $n$ \\
\ precision $Q$: $n \times n$ symmetric matrix with $l$ sub-diagonals  \\
\ parameter $\theta$ and density $p(\theta)$}
\Result{\   Log likelihood $\;\log p(v, \theta, Y)$}
\begin{algorithmic}
\STATE \% Cholesky factor of $Q$
\STATE $L_Q \leftarrow C(Q)$  \% $n \times n$, $l$ sub-diagonals
\STATE \% Construct observation sample
\STATE $F \leftarrow \opS(L_Q^T,v) $
\STATE \% Compute log-likelihood
\STATE $\log p(v, \theta, Y) \leftarrow
	\log\,p(v) +  \log\,p(\theta) + \sum_{i=1}^n \log\,p(y_i|\theta, F_i).$
\RETURN $\log p(v, \theta, Y)$
\end{algorithmic}
\end{algorithm}

\subsection{Variational Inference Objective}

In section \ref{subsec:vi}, we introduced a novel algorithm to perform variational inference in GMRF models using a precision parameterised Gaussian variational distribution. Algorithm \ref{algo:vi} provides some details of how we use our operators for banded matrices to evaluate the variational lower bound to the marginal likelihood.

\begin{algorithm}[h!]
   \caption{Variational Inference objective evaluation.}
   \label{algo:vi}
\SetKwInOut{Input}{Input}  
\SetKwInOut{Result}{Output}  
\Input{\ prior parameters:\\
\ $m_p$: vector of length $\n$ \\
\ $L_p$: $\n \times \n$ lower triangular matrix with $l_p$ sub-diagonals \\
\ variational parameters:\\
\ $m_q$: vector of length $\n$ \\
\ $L_q$: $\n \times \n$ lower triangular matrix with $l_q$ sub-diagonals\\
\ observations $Y$: vector of length $\n$}
\Result{\  variational objective $\cL$}

\begin{algorithmic}
\STATE \% Compute marginal variance \\
\STATE $\sigma^2_F \leftarrow diag [\opI(L_F)]$.\\
\STATE \% Evaluate KL divergence \\
\STATE $KL \leftarrow \frac{1}{2}  \left( \mathrm{tr} (  \opI(L_q) \opP(L_p,L_p^T) ) + 2 \sum_i (\log [L_q]_{ii} - \log [L_p]_{ii})
               +  ||\opP(L_p^T,  m_p -m_q)||^2  - \n   \right)$\\
\STATE \% Evaluate variational expectations \\ 
\STATE $VE \leftarrow \sum_{i=1}^n \EE_{q(F_i)=\NN(\mu_{F,i}, \sigma^2_{F,i})} \log p(Y_i|F_i)$\\
\STATE \% Return variational lower bound\\ 
\STATE $\cL \leftarrow VE - KL$\\
\RETURN $\cL$
\end{algorithmic}
\end{algorithm}

\subsection{Marginal Likelihood}

In section \ref{subsec:loglik}, we explained how to efficiently compute marginal likelihood compuation in conjugate GMRFs with partial observations. In Algorithm \ref{algo:loglik}, we give further details about how we use our operators for banded matrices to achieve these efficient compuations.

The generative model is the following: latent vector $f$ has a zero mean multivariate Gaussian prior with precision $Q$, observation model is additive white noise with variance $\tau^2$.

\begin{algorithm}[h!]
   \caption{Marginal Likelihood}
   \label{algo:loglik}
\SetKwInOut{Input}{Input}  
\SetKwInOut{Result}{Output}  
\Input{\ parameter $\theta$\\
\ prior precision $Q$: $n \times n$ symmetric matrix with $l$ sub-diagonals \\
\ observations $Y$: vector of length $n$\\
\ indicator matrix $E$: matrix of size $\n \times \N$}
\Result{\   $\log\,p(Y| \theta)$ }
\begin{algorithmic}
\STATE \% Cholesky factorization of the posterior precision
\STATE $L \leftarrow \opC(Q  + \tau^{-2} E^\T E)$ \% note that $E^\T E$ is diagonal
\STATE \% Cholesky factorization of the prior precision
\STATE $L' \leftarrow \opC(Q)$
\STATE \% Evaluation of the Marginal likelihood
\STATE $R \leftarrow \opS (L, E^{T} Y)$
\STATE $\log\,p(Y| \theta) \leftarrow  -\frac{\n}{2} \log(2 \pi) - \log |L| + \log |L'| - \frac{1}{2 \tau^2} Y^\T Y + \frac{1}{2 \tau^4} R^\T R$
\RETURN $\log\,p(Y| \theta)$
\end{algorithmic}
\end{algorithm}

\subsection{Classical Kalman filter recursions}
\label{sec:appendix:classicKalman}
In section \ref{subsec:kalman}, we compared our implementation of Gaussian process regression (for kernels have a state-space representation) using our operators to traditional implementations using Kalman filtering. In this section and summarized in Algorithm \ref{algo:classic_kalman} are derivations of the classic Kalman filter recursions.\\

We implement the filtering recursions for a linear state-space model with state dimension $d$ and observation dimension $e$ defined as 
\[
F_t \sim \NN(F_t; A_t F_{t-1}+b_t; Q_t), \quad Y_t \sim \NN(Y_t; H F_t +c, R),
\]
with $F_0 \sim \NN(F_0;\mu_0, \Sigma_0)$. For a sequence of $T$ observations stacked into $Y^T = [Y_1^\T,\dots, Y_T^\T]\in \RR^{Te}$, the aim is to evaluate $p(Y)$ which can be expressed in terms of conditionals densities $p(Y) = \prod_t p(Y_t|Y_{1\dots t-1})$. These conditional densities can be obtained after one pass of the Kalman filtering recursions as follows:

Define $p(F_t|Y_{1\dots t-1}) = \NN(F_t; \mu_t, \Sigma_t )$. We update the belief on $F_t$ after observing $Y_t$:
\begin{align*}
p(F_{t} |Y_{1\dots t}) &= p(F_{t} |Y_{1\dots t-1}, Y_t)\\
 &\propto p(Y_t |F_{t}, Y_{1\dots t-1}) p(F_{t}| Y_{1\dots t-1})\\
 &\propto p(Y_t |F_{t}) p(F_{t}| Y_{1\dots t-1})\\
 &= \NN( F_t; \bar{\mu}_t, \bar{\Sigma}_t)
\end{align*}
with 
\begin{align*}
\bar{\Sigma}_{t}  &= (\Sigma_t^{-1} + H^\T R^{-1} H)^{-1} \\
&= \Sigma_t - \Sigma_t H^\T (R + H \Sigma_t H^\T)^{-1} H \Sigma_t \\
&= (I  - K_t H)\Sigma_t ,\quad K_t=\Sigma_t H^\T (R + H \Sigma_t H^\T)^{-1}  \\
\bar{\mu}_{t} &= \bar{\Sigma}_{t}(\Sigma_t^{-1}\mu_t + H^\T R^{-1} (Y_t-c))\\
 &= (I  - K_t H)\mu_t + K_t(Y_t-c).
\end{align*}
$K_t$ corresponds to the Kalman gain. Then we predict one step ahead:
\begin{align*}
p(F_{t+1}|Y_{1\dots t}) &=  \int\, d F_t p(F_{t+1}|F_{t})p(F_{t} |Y_{1\dots t}) \\
&= \NN(F_t; \mu_{t+1}, \Sigma_{t+1} ),
\end{align*}
so 
\begin{align*}
	\Sigma_{t+1}  &= A_{t+1} \bar{\Sigma}_{t} A_{t+1}^\T + Q_{t+1} \\
	\mu_{t+1} &= A_{t+1}\bar{\mu}_{t} + b_{t+1}.
\end{align*}
In the end, for the marginal likelihood computation, we need:
\begin{align*}
p(Y_{t}|Y_{1 \dots t-1}) &= \int\,
 dF_{t} p(Y_{t}|F_{t})p(F_{t}|Y_{1\dots t-1})\\
&=  \NN( Y_t; H\mu_t, H\Sigma_t H^\T + R).
\end{align*}

\begin{algorithm}[h!]
 \caption{Kalman Filtering}
 \label{algo:classic_kalman}
 \SetKwInOut{Input}{Input}  
\SetKwInOut{Result}{Output}  
\Input{\ parameters $\theta=[A_t, b_t, Q_t, H, c, R]$, Observations $Y$}
\Result{\   $\log\,p(Y| \theta)$ }
\begin{algorithmic}
\STATE $L \leftarrow 0$
\FOR {$t = 1\dots T$ }
\STATE \% incorporation of observation $Y_t$
\STATE $K_t \leftarrow \Sigma_t H^\T (R + H \Sigma_t H^\T)^{-1}$
\STATE $\bar{\Sigma}_{t}  \leftarrow (I  - K_t H)\Sigma_t$ \\
\STATE $\bar{\mu}_{t} \leftarrow (I  - K_t H)\mu_t + K_t(Y_t-c)$ \\
\STATE \% prediction one step ahead
	\STATE $\Sigma_{t+1}  \leftarrow A_{t+1} \bar{\Sigma}_{t} A_{t+1}^\T + Q_{t+1}$ \\
	\STATE $\mu_{t+1} \leftarrow A_{t+1} \bar{\mu}_{t} + b_{t+1}$
\STATE \% evaluation of the log conditional $\log \,p(Y_{t}|Y_{1\dots t-1})$
\STATE $L \leftarrow L +  \log \NN( Y_t; H\mu_t, H\Sigma_t H^\T + R)$
\ENDFOR
\RETURN $L$
\end{algorithmic}
\end{algorithm}

\subsection{Gaussian process regressiong with banded precision}

Our implementation of Gaussian process regression, for the experiment of section \ref{subsec:kalman}, treats the chain of latent states $F^\T = [F_0^\T,...,F_T^\T]\in \RR^{(T+1)d}$ and observations $Y^\T = [Y_1^\T,...,Y_T^\T]\in \RR^{Te}$ as a big multivariate normal distribution $p(Y,F)$ that factorises into a prior and a likelihood as $p(Y,F)=p(Y|F)p(F)$. Both terms can be seen as multivariate normal densities on $F$, albeit a possibly degenerate one for $p(Y|F)$.
They can be expressed in terms of their natural parameters:
\begin{align*}
p(F) &= \NN(F; \eta_{\prior}, \Lambda_{\prior})\\
p(Y|F) &= \NN(F; \eta_{\lik}, \Lambda_{\lik}),
\end{align*}
where only $\eta_{\lik}$ depends on the data $Y$.

The density of a multivariate normal distribution with natural parameterisation $\eta, \Lambda$ is: 
\begin{align*}
\NN(F; \eta, \Lambda) &= \exp\left( a(\eta, \Lambda) + \eta^\T F + \frac{1}{2}F^\T \Lambda F \right)\\
a(\eta, \Lambda) &= -\frac{1}{2} \left(-\log |\Lambda| + Td \log 2\pi  + \eta^\T \Lambda^{-1} \eta \right) 
\end{align*}

Hence the joint over latent $F$ and observations $Y$ is
\begin{align*}
p(F,Y) &= \NN(F, \eta_{\prior}, \Lambda_{\prior}) \NN(F, \eta_{\lik}, \Lambda_{\lik})\\
&=  \exp\left( a(\Lambda_{\prior},\eta_{\prior}) + a(\Lambda_{\lik},\eta_{\lik}) - a(\Lambda_{\post}, \eta_{\post}) \right)  \NN(F; \eta_{\post}, \Lambda_{\post})
\end{align*}
with $(\eta_{\post},\Lambda_{\post}) = (\eta_{\prior},\Lambda_{\prior}) + (\eta_{\lik},\Lambda_{\lik})$ the parameters of the Gaussian posterior on $F$.

Marginalizing over $F$ provides the log marginal 
\begin{align*}
\log p(Y) &=  a(\Lambda_{\prior},\eta_{\prior}) + a(\Lambda_{\lik},\eta_{\lik}) - a(\Lambda_{\post}, \eta_{\post}).
\end{align*}

Due to the structure of the model, the prior and likelihood terms can be further simplified: \\
Since $p(F) = p(F_0) \prod_t p(F_t|F_{t-1})$, we have $ |\Lambda_{\prior}| = |\Sigma_0|^{-1}\Pi_t|Q_t|^{-1}$.\\
 Since $p(Y|F) =\prod_t p(Y_t|F_t)$ and noting $p(Y_t|F_t)=\NN(F_t; \eeta_t, R)$, we have $|\Lambda_{\lik}| = |R|^{-T}$ and $\eta_{\lik}^\T \Lambda_{\lik}^{-1} \eta_{\lik}= \sum_t \eeta_t^\T R \eeta_t$.

All the other log determinants and quadratic terms are expressed from the Cholesky factor $C_j$ of $\Lambda_j$ as $ \log |\Lambda_j| =  2\log |C_j| = 2\sum_i \log [C_j]_{ii}$ and $\eta_j^\T \Lambda_j^{-1} \eta_j = ||C_j^{-1} \eta_j||^2$.

So we can rewrite the likelihood as 
\begin{align*}
\log p(Y) &=  \frac{1}{2}  \left( Td\log 2\pi - \log |\Sigma_0 R^T \Pi_t Q_t| + 2\log |C_{\post}| \right) \\
 &+ \frac{1}{2} ( ||C_{\prior}^{-1} \eta_{\prior} ||^2 + \sum_t \eeta_t^\T R \eeta_t - ||C_{\post}^{-1} \eta_{\post} ||^2 )
\end{align*}

\begin{algorithm}[h!]
 \caption{Gaussian process regression (using banded matrices)}
 \label{algo}
 \SetKwInOut{Input}{Input}  
\SetKwInOut{Result}{Output}  
\Input{\ parameters $\theta=[A_{1\dots T}, b, Q_{1\dots T}, H, c, R]$,\\
\ observation $Y$: vector of length $\n$}
\Result{\   $\log\,p(Y| \theta)$ }
\begin{algorithmic}
\STATE \% building the prior and likelihood natural parameters [see Grigorievskiy et al 2017]
\STATE $\eta_{\prior}, \Lambda_{\prior} \leftarrow f(A_{1\dots T},b,Q_{1\dots T})$
\STATE $\eta_{\lik}, \Lambda_{\lik} \leftarrow g(H,c,R,Y)$ 
\STATE \% Cholesky factorization of the prior and posterior precisions 
\STATE $C_{\prior} \leftarrow  \opC(\Lambda_{\prior} )$
\STATE $C_{\post} \leftarrow  \opC(\Lambda_{\prior} + \Lambda_{\lik})$
\STATE \% evaluation of the log-likelihood
\STATE $\log p(Y) \leftarrow  \frac{1}{2} \left( Td  \log 2\pi
 -   \log |\Sigma_0 R^T \Pi_tQ_t| - 2\log |C_{\post}|  
 +   ||C_{\prior}^{-1} \eta_{\prior} ||^2 + \sum_t \eeta_t^\T R \eeta_t - ||C_{\post}^{-1} \eta_{\post} ||^2 \right)$
\RETURN $L$
\end{algorithmic}
\end{algorithm}

The bottleneck in computational terms is the Cholesky factorization which scales as $O(Tl^3)$, because $\Lambda_{post}$ is a banded precision of length $Tl$ and bandwidth $2l$.

\subsection{Gaussian process regression}

In section \ref{subsec:kalman}, we compare different implementations of Gaussian process regression to the naive one that builds a dense covariance matrix $K$ for the latent.
For this latter case, the marginal likelihood is evaluated as: 
\[
\log \,p(Y|\theta) = -\frac{1}{2}Y^\T(K + \sigma^2 I)^{-1}Y -\frac{1}{2} \log |K + \sigma^2 I| - \frac{N}{2} \log 2\pi 
\]
Algorithm \ref{algo:gpr} provides details on how we implemented it in practice.

\begin{algorithm}[h!]
   \caption{Gaussian process regression Likelihood}
   \label{algo:gpr}
\SetKwInOut{Input}{Input}  
\SetKwInOut{Result}{Output}  
\Input{\ parameter $\theta$\\
\ observation $Y$: vector of length $\n$}
\Result{\   $\log\,p(Y| \theta)$ }
\begin{algorithmic}
\STATE \% Cholesky factorization of the posterior covariance
\STATE $L \leftarrow \text{Cholesky}(K + \sigma^2 I)$
\STATE \% marginal likelihood evaluation
\STATE $\log\,p(Y| \theta) \leftarrow -\frac{1}{2}||L^{-1} Y||^2 - \sum_i \log|L_{ii}| - \frac{N}{2} \log 2\pi$
\RETURN $\log\,p(Y| \theta)$
\end{algorithmic}
\end{algorithm}